\documentclass{IOS-Book-Article}
\usepackage{mathptmx}
\usepackage{amsmath}
\usepackage{amsfonts}
\usepackage{amssymb}
\usepackage{graphicx}
\usepackage{pstricks}
\usepackage{verbatim}
\usepackage{url}
\usepackage{makeidx}  
\usepackage{amsmath}
\usepackage{graphicx} 
\usepackage{hyperref} 
\hypersetup{colorlinks=true,urlcolor=black,citecolor=black,linkcolor=black}
\usepackage{multirow}
\usepackage{enumerate}
\usepackage{wrapfig}
\usepackage{float}
\usepackage{floatflt}
\usepackage{caption}
\usepackage{algorithm}
\usepackage{tikz}
\usetikzlibrary{arrows,shapes,snakes,matrix,decorations.pathmorphing,backgrounds,fit,positioning,shapes.symbols,chains}
\usepackage{amsthm}
\usepackage{algpseudocode}
\algnewcommand\algorithmicforeach{\textbf{for each}}
\algdef{S}[FOR]{ForEach}[1]{\algorithmicforeach\ #1\ \algorithmicdo}
\def\hb{\hbox to 10.7 cm{}}

\begin{document}

\pagestyle{headings}
\def\thepage{}

\begin{frontmatter}              

\title{Amnestic Forgery: an Ontology of Conceptual Metaphors}
\runningtitle{Amnestic Forgery}

\author[A,B]{\fnms{Aldo} \snm{Gangemi}} 
\author[B]{\fnms{Mehwish} \snm{Alam}} 
\author[B]{\fnms{Valentina} \snm{Presutti}} 

\runningauthor{A. Gangemi et al.}
\address[A]{\tiny FICLIT, University of Bologna, Italy}
\address[B]{\tiny Semantic Technology Lab, ISTC-CNR, Rome, Italy}

\begin{abstract}
This paper presents Amnestic Forgery, an ontology for metaphor semantics, based on MetaNet, which is inspired by the theory of Conceptual Metaphor. Amnestic Forgery reuses and extends the Framester schema, as an ideal ontology design framework to deal with both semiotic and referential aspects of frames, roles, mappings, and eventually blending. The description of the resource is supplied by a discussion of its applications, with examples taken from metaphor generation, and the referential problems of metaphoric mappings. Both schema and data are available from the Framester SPARQL endpoint. 
\end{abstract}

\end{frontmatter}

\section{Introduction}
\label{intro}
A metaphor is a cognitive operation involving usage of natural language and cross-domain conceptual mapping. Its ontological interest is in principle purely cognitive and linguistic, i.e. how to model what humans do when a metaphorical mapping is activated by speech or text. 
However, ontology-based extraction and representation of knowledge needs to make the semantics of natural language explicit, establishing the  referential aspects of a natural language construction as used in dialogues, descriptions, memorisation, fiction, poetry, instructions, emotional expression, etc. -- in other words, for any function of language \cite{jakobson}.

In this paper we took the bull by the horns, and straightforwardly designed an OWL ontology for metaphors, mappings, blending, etc., and populated it with data from Berkeley's MetaNet \cite{W15-1405}. MetaNet is the reference repository of conceptual metaphors, developed as a Semantic Wiki\footnote{\url{https://metaphor.icsi.berkeley.edu/pub/en/}}, maintained through collaborative editing by multiple conceptual metaphor experts (cf. Sect. \ref{sec:metanet}).

The new ontology is called Amnestic Forgery\footnote{This is a recursive name, since FORGERY IS AMNESIA is a new metaphor generated by means of the ontology itself, cf. Sect. \ref{generating}.}. We deploy it as an extension of Framester\footnote{\url{http://etna.istc.cnr.it/framester2/sparql}} 
\cite{Gangemi:2016:FWC:3092960.3092977} a knowledge graph represented as Linked Open Data (LOD), which integrates heterogeneous linguistic resources (OntoWordNet \cite{fellbaum98wordnet}, VerbNet \cite{KipperDP00}, FrameNet-OWL \cite{shortBaker1998,Nuzzolese:2011:GLL:1999676.1999685}, BabelNet \cite{Navigli:2012:BAC:2397213.2397579}, etc.), factual datasets (DBpedia \cite{LehmannEtAl09}, YAGO \cite{Suchanek:2008:YLO:1412759.1412998}, etc.), and foundational ontologies, by providing them a unified formal semantics.
We give a brief introduction to MetaNet and FrameNet in Sect. \ref{sec:metanet} and then discuss the problem of creating a metaphor representation (Sect. \ref{metaphor}), extracting MetaNet data and their schema, and aligning them to Framester (Sect. \ref{amnestic}), as well as envisaging multiple use cases with practical examples (Sect. \ref{generating}). We also discuss referential problems of metaphorically filtered situations (Sect. \ref{referential}). We complement the paper with a survey of computational metaphor studies (Sect. \ref{related}), and conclusions.

\section{MetaNet and FrameNet}\label{sec:metanet}

MetaNet~\cite{W15-1405} is a repository of manually curated metaphors. It has its roots dug into the linguistic frames as available on FrameNet~\cite{shortBaker1998}. FrameNet is a resource containing conceptual frames, where each frame consists of a description of a situation as denoted by a text; e.g., the frame \texttt{Studying} depicts the situation where a \texttt{student} is performing an act of studying in some \texttt{institution} (student and institution are the ``semantic roles'' used to encode a studying situation).

\begin{figure}
 \begin{center}
\begin{tabular}{c}

\begin{tikzpicture}[scale=0.6, every node/.style={scale=0.75}]
     \tikzstyle{ann} = [draw=none,fill=none,right]
    
     \matrix[nodes={draw},
         row sep=0.4cm,column sep=0.6cm] {
&&\node[fill=none] (a) {$Thinking~Is~Linguistic~Activity$};&& \\
&\node[fill=none] (c) {$Simple~Ideas~Are~Words$};&\node[fill=none] (b) {$Memorization~Is~Writing$};&\node[fill=none] (d) {$Thinking~Is~Speaking$};&\\
&&&&\\
&\node[ellipse,fill=none] (e) {$Writing$};&&\node[ellipse,fill=none] (f) {$Memorization$};&\\
     };
     
      \draw [->] (b)  to node [above] {}  (a);
      \draw [->] (c)  to node [above] {}  (a);
      \draw [->] (d)  to node [above] {}  (a);
      \draw [-] (e)  to node [left] {Source Frame}  (b);
      \draw [-] (f)  to node [right] {Target Frame} (b);
 \end{tikzpicture}
 
  \end{tabular}
  \end{center}
    \caption{An example of network of metaphors contained in MetaNet. The arrows  represent the \emph{Inheritance} relations as defined in MetaNet.}
     \label{fig:metanet}
\end{figure}
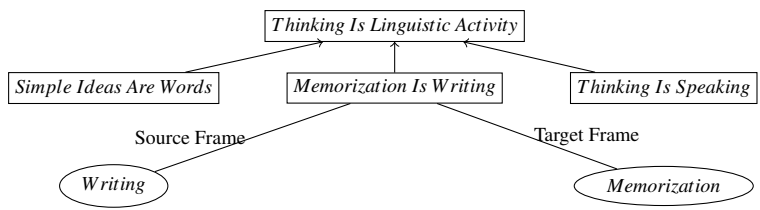

In \cite{W15-1405}, the authors describe an automated system for extracting metaphors using the information from a manually built metaphor repository. The repository contains metaphors along with their conceptual frames, metaphor constructions, and metaphoric relational patterns. Figure~\ref{fig:metanet} shows an example of a metaphor {\bf Memorization is Writing} from MetaNet, which is related to another metaphor {\bf Thinking is Linguistic Activity}. The other neighbouring metaphors such as {\bf Simple Ideas Are Words} and {\bf Thinking Is Speaking} are also connected to the same generic metaphor. The ellipses represent the source and the target frames of one of the metaphors i.e., {\bf Memorization Is Writing}. {\emph ``Memorization"} frame already exists in FrameNet while {\emph ``Writing"} frame is a MetaNet specific frames.

In Sect. \ref{amnestic} we describe how the informal encoding of MetaNet has been extracted, refactored, and integrated into the Framester knowledge graph.

\section{Representing metaphors and blending in linguistic-factual semantics}
\label{metaphor}
The range of metaphorically-laden linguistic constructions is large. In order to focus on the semantic aspects of metaphors, we mostly provide examples from a unique linguistic construction: the adjective-noun phrase modification. As we reported in \cite{DBLP:conf/fois/GangemiNPR16}, when an adjective modifies a noun, we can represent that modification as a frame composition. The composition can have a conservative effect upon the compositional semantics of the discourse (as in the case of a specification), or can act as a non-conservative extension, as in the case of metaphors, and blending in general \cite{fauconnier2002wwt}.

Examples for non-metaphorical adjectival modification include global intersective composition (\emph{American woman}), local attributional composition (\emph{skillful woman}), attitude-laden frame construction (\emph{alleged woman}), or ``privative'' novel frame creation (\emph{stony woman}). This can be easily understood when considering a fixed \texttt{Woman} frame, and composing it with a \texttt{Nationality} frame vs. a \texttt{Competence} frame vs. an \texttt{AttitudeTowards} frame, vs. a \texttt{ConstitutingMaterial} frame. These four composition types work in different ways, but what they basically do is to establish a new referential frame, which, in the case of conservative composition, (1) inherits from the core one (\texttt{Nationality+Woman}, \texttt{Competence+Woman}), or announces a meta-level predication over the core frame (\texttt{AttitudeTowards(Woman)}), or, in the case on non-conservative composition, (2) provides instructions to create a blending (\texttt{ConstitutingMaterial+Woman}), in which a new frame emerges that does not inherit from the core one, rather it reuses part of the roles in the core frame (\texttt{Woman}), while substituting others with the blended frame (\texttt{ConstitutingMaterial}).

When dealing with adjectives that are used metaphorically (e.g. \emph{stony woman} in the sense of a woman that does not show feelings or sympathy), classical theories of metaphor (from structure mapping \cite{gent83} to embodied conceptual metaphor \cite{LakoffJohnson80}) tell us that two frames are blended in more or less fixed ways, where e.g. certain roles from the \texttt{Feeling} frame associated with a target frame \emph{Woman} are substituted by roles from a source frame (\emph{ConstitutingMaterial}).
Blending is therefore more general than the case with privative frame blending, and covers metaphor as well. In practice, blending operates not only when inheritance from the core frame is prevented (the privative case), but also when an internal role of the core frame is substituted by a role from the modifying one.

Metaphoricity (the property of having a metaphorical interpretation) is sensible to modification structures in language \cite{Morzycki15}, so that e.g. adjective-noun constructions tend to have a double interpretation when the two frames seem to be incompatible. In \emph{stony woman}, we have specific frame incompatibility (stone as a material is incompatible with roles from the \texttt{Woman} frame), but depending on the modification semantics adopted by the interpreter, the referential semantics changes completely. In the privative case, the head of the construction (\emph{woman}) is interpreted as a fake woman made of stone, e.g. a statue, while in the metaphoric case, the \emph{stony} modifier is interpreted by blending the \texttt{ConstitutingMaterial} and \texttt{Feeling} frames.

Distinguishing between privative and metaphoric interpretation of apparently incompatible modification is by no means trivial, and requires a larger frame composition, supporting the need for deep knowledge extraction in order to approximate human-level Natural Language Understanding (NLU). For example, in sentence \ref{giza}:
\begin{equation} \label{giza}
\text{\parbox{.85\textwidth}{The Giza sphinx is a chimeric stony woman}}
\end{equation}

the type of the header \emph{woman} is not Organism, but Statue, and lets us adopt the privative interpretation. On the contrary, in the sentence \ref{creeps}:
\begin{equation} \label{creeps}
\text{\parbox{.85\textwidth}{The few words from that stony woman gave me the creeps}}
\end{equation}

the type of the header is not changed, and a metaphoric interpretation must be applied. Anyway, more intermediate cases can be imagined: if deep extraction from full discourse parsing is needed to distinguish between these clear examples, much more background knowledge would be needed to achieve accuracy in arbitrary sentences.
For this very reason, we propose to integrate metaphoric and blending knowledge into a huge graph of linguistic and factual knowledge, Framester \cite{Gangemi:2016:FWC:3092960.3092977}.

Conceptual Blending theory is based on the procedure of abstracting a generic space out of two input spaces, and then generating a new blended space.
When used with reference to Conceptual Metaphor theory, this corresponds to the abstraction of a new frame that is more generic than either the target or source frames, and the creation of a new blended frame that reuses roles from any of the two input frames.
Both the abstraction and the creation processes can be very difficult or unpredictable, as expected in a creative process that involves many different parameters from language, culture, physical environment, affordances, etc.

Our hypothesis of a framal semantics for adjective modification as proposed in \cite{DBLP:conf/fois/GangemiNPR16} is here extended to cover all sorts of metaphorically-based modification: blending is a ``warping'' composition, which obliges the interpreter to construct a new acceptable reference frame by using roles taken from two different frames.

Empirically, previous work on framal adjective semantics \cite{DBLP:conf/fois/GangemiNPR16} has been extended by generalising the frame semantics underlying any piece of linguistic content or data, as formalised in the Framester knowledge graph (\cite{Gangemi:2016:FWC:3092960.3092977}), which adds frame annotations to millions of words, word senses, synsets, individuals, classes, relations, etc. Recently, we have converted the MetaNet wiki information as a RDF knowledge graph with an OWL ontology compatible to the Framester schema, linked MetaNet frames to existing Framester frames, and added the resource to Framester, in order to make it queryable with ordinary APIs for computational experiments.
In this paper, we report about the conversion and integration of MetaNet into Framester, and the ontology that resulted from it, showing some knowledge graph operations on Framester, which allow to generate candidate metaphor extensions. 


\section{Amnestic Forgery: an ontology for MetaNet and beyond}
\label{amnestic}
Framester ontology is based on Descriptions and Situations (D\&S) \cite{DBLP:journals/aamas/Gangemi08, ontolexschema}, a flexible ontology pattern framework that can be used with any reasoning pipeline in order to perform classification, partial matching, diagnosis, abstraction, or construction operations between a theoretical (Description, Frame) structure, and a factual (Situation, Frame Occurrence) structure.

The D\&S formal framework is perfectly echoed in the (informal) explanation of the FrameNet core schema: ``For example, the \textbf{Apply heat} frame describes a common situation involving a Cook, some Food, and a Heating Instrument, and is evoked by words such as bake, blanch, boil, broil, brown, simmer, steam, etc. We call these roles frame elements and the frame-evoking words are lexical units in the \textbf{Apply heat} frame''.

Once formalised in D\&S, the FrameNet implicit schema becomes a semiotic passepartout: a frame $f$, as a description, can be the reification of any relation $\rho$ with arbitrarily variable arity, a frame element $fe$ is a binary projection of $\rho$, and a lexical unit $lu$ of $f$ is a symbol, for which $\rho$ (and its reified counterpart $f$), and its projections $fe_{1...n}$ act as intensional interpretations. A ``common situation'' $s$ described by $f$ is the extensional interpretation (aka \emph{denotation}) of $lu$, whose intension is $f$.

Since D\&S allows descriptions to be composed of unary concepts, and situations to be composed of arbitrary entities, the game becomes more interesting, and enables the formal representation for the dependency of $fe$ on $f$, and the formal construction of $s$ out of arbitrary entities $e_{1...n}$ corresponding to the projections $fe_{1...n}$ of $f$. An extensive explanation of the FrameNet-OWL resource designed according to D\&S is presented in \cite{Nuzzolese:2011:GLL:1999676.1999685}.

Later, this approach to abridge semiotic and model-theoretical representation of frame semantics has been broadened in order to encompass any linguistic or factual resource, and opened the way to Framester \cite{Gangemi:2016:FWC:3092960.3092977}, a large knowledge graph containing more than 50 million triples linking millions of linguistic, conceptual, or real world entities. In Framester, D\&S-based frame semantics allows to reduce the heterogeneity of the targeted resources as follows.

Framester implements a dual frame semantics by means of OWL2 \cite{owl2-overview} punning, so that each instance of the \texttt{Frame} class is also a subclass of the \texttt{FrameOccurrence} class.

Furthermore, the \emph{FrameProjection} class allows to integrate any predicate defined either intensionally or extensionally in ontologies, lexical resources, or other vocabularies or web formats, For example, the FrameNet frame \texttt{Activity\_start} as well as the VerbNet verb class \texttt{verbclass-begin-55.1-1} are linked as intensionally equivalent to the Framester frame \texttt{ActivityStart}, while the synset \texttt{synset-newcomer-noun-1} from WordNet, which is intensionally mapped to FrameNet \texttt{Activity\_start}, is  extensionally represented as a class of newcoming entities, and linked as a \emph{unary projection} of a Framester class of newcoming situations, \texttt{Newcomer.n.1}, which on its turn is represented as a subclass of \texttt{ActivityStart}.

An \texttt{InternalBinaryProjection} of a frame, such as semantic roles from FrameNet, VerbNet, PropBank, the Preposition Project, etc., as well as properties from factual resources as e.g. the DBpedia Ontology, are aligned to generic roles in Framester, leading to better interoperability when extensionally represented as internal binary projections of a frame.

An \texttt{ExternalBinaryProjection} of a frame (e.g. a relation between the agent and the location of a situation for the frame \texttt{ActivityStart}) can also be generalised by mapping them to pairs of internal binary projections.

Finally, individuals classified with types that can be disambiguated as unary projections of a frame (e.g. ``a newcomer''), can be formalised as potential evokers of that frame in the context of a discourse.

In D\&S, higher-level descriptions can also be defined, e.g. for meta-norms that describe priority between other norms. Accordingly, frame composition (see Sect. \ref{metaphor} can be formalised in D\&S as a higher-level description that creates a new frame by \emph{merging} two frames (globally or locally, or with modifications such as epistemic or deontic attitudes), or by \emph{blending} two frames.
With merging, the higher-level description is quite simple, since it uses the value of a modifying frame as a role filler in the modified frame, so generating a specification.
With blending, the representation is more complex: a higher-level description that incorporates operators to abstract two input descriptions, and to create a new blended frame.
The typical outcome of reasoning with a blending can be a metaphor, as in the case of \emph{stony woman} (in the behavioral sense), or a type selection, as with \emph{stony woman} (in the material sense).
A metaphor itself is a kind of description, which incorporates roles for two more descriptions (the source and target frames), as well as mapping rules between the respective roles.

Equipped with this intuition, originally sketched in \cite{ontolexschema}, and already used in several projects to formalise linguistic resources (OWL versions of FrameNet, VerbNet, PropBank, etc.), we have the possibility to reuse the largest D\&S-based knowledge base, Framester \cite{Gangemi:2016:FWC:3092960.3092977}.

The design approach taken to formalise MetaNet is to use D\&S in order to extract and formalise a metaphor, then to extract data and formally represent them according to that schema, and finally to align the schema and data to elements in the Framester knowledge graph. 

\begin{figure}
\centering
  \includegraphics[width=12.5cm,keepaspectratio=true]{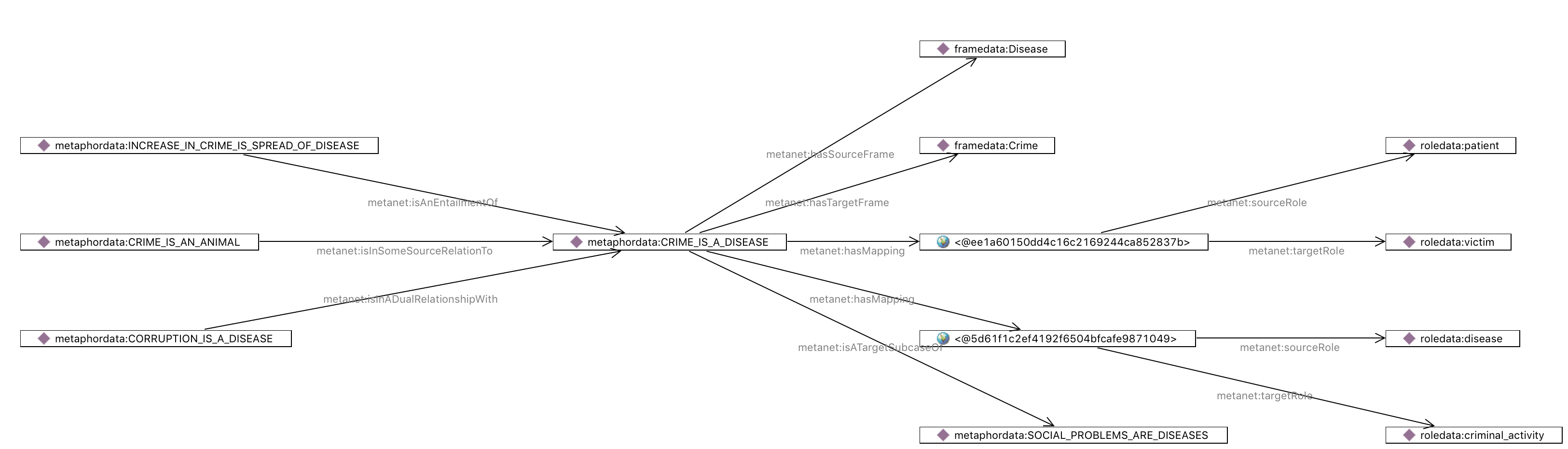}
 \caption{The subgraph for the metaphor CRIME\_IS\_A \_DISEASE.}
 \label{metaphorexample}
\end{figure}

We have firstly scraped tabular data from the MetaNet wiki\footnote{The MetaNet wiki is a SemanticMediaWiki instance, but its data querying facility is not accessible.}, and we have designed a preliminary MetaNet schema that catches the intended meaning of the interface used to populate the MetaNet wiki.
Secondly, we have refactored the extracted data according to this preliminary schema, and fine-tuned it against features deriving from the data entry variety in the wiki. The result is a refined schema, the Amnestic Forgery ontology, and its MetaNet data.
Fig. \ref{metaphorexample} depicts a subgraph of MetaNet for the metaphor CRIME\_IS\_A \_DISEASE. The subgraph contains examples of the core relations in MetaNet, linking metaphors to their source and target frames, their role mappings, entailments, and possibly other more vague relations contributed by the users of the wiki.

\begin{figure}
\centering
  \includegraphics[width=12.5cm,keepaspectratio=true]{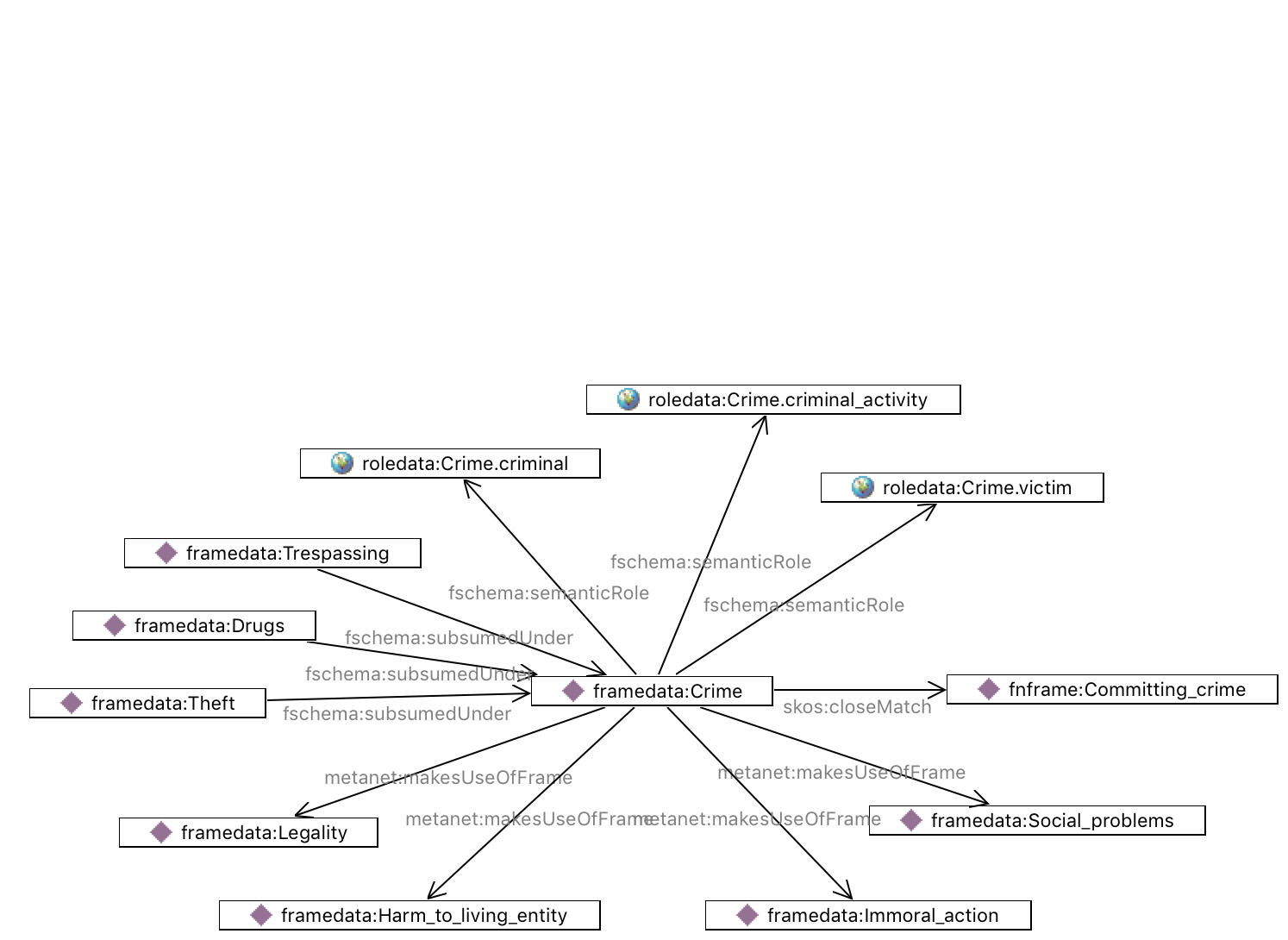}
 \caption{The subgraph for the frame \texttt{Crime}.}
 \label{frameexample}
\end{figure}

Fig. \ref{frameexample} depicts a subgraph of MetaNet for the frame \texttt{Crime}. The subgraph contains examples of the core relations in MetaNet, linking frames to their roles, their more specific frames, their alignments to frame from other resources, and other frames bearing dependency relations.

Notice that the subgraphs showed here are examples of the intensional view of Amnestic Forgery, while the extensional view adds axioms that have metaphors and frames as classes of situations, and relations to roles as OWL restrictions (quantified clauses as superclasses of frames).

\begin{figure}
\centering
  \includegraphics[width=12.5cm,keepaspectratio=true]{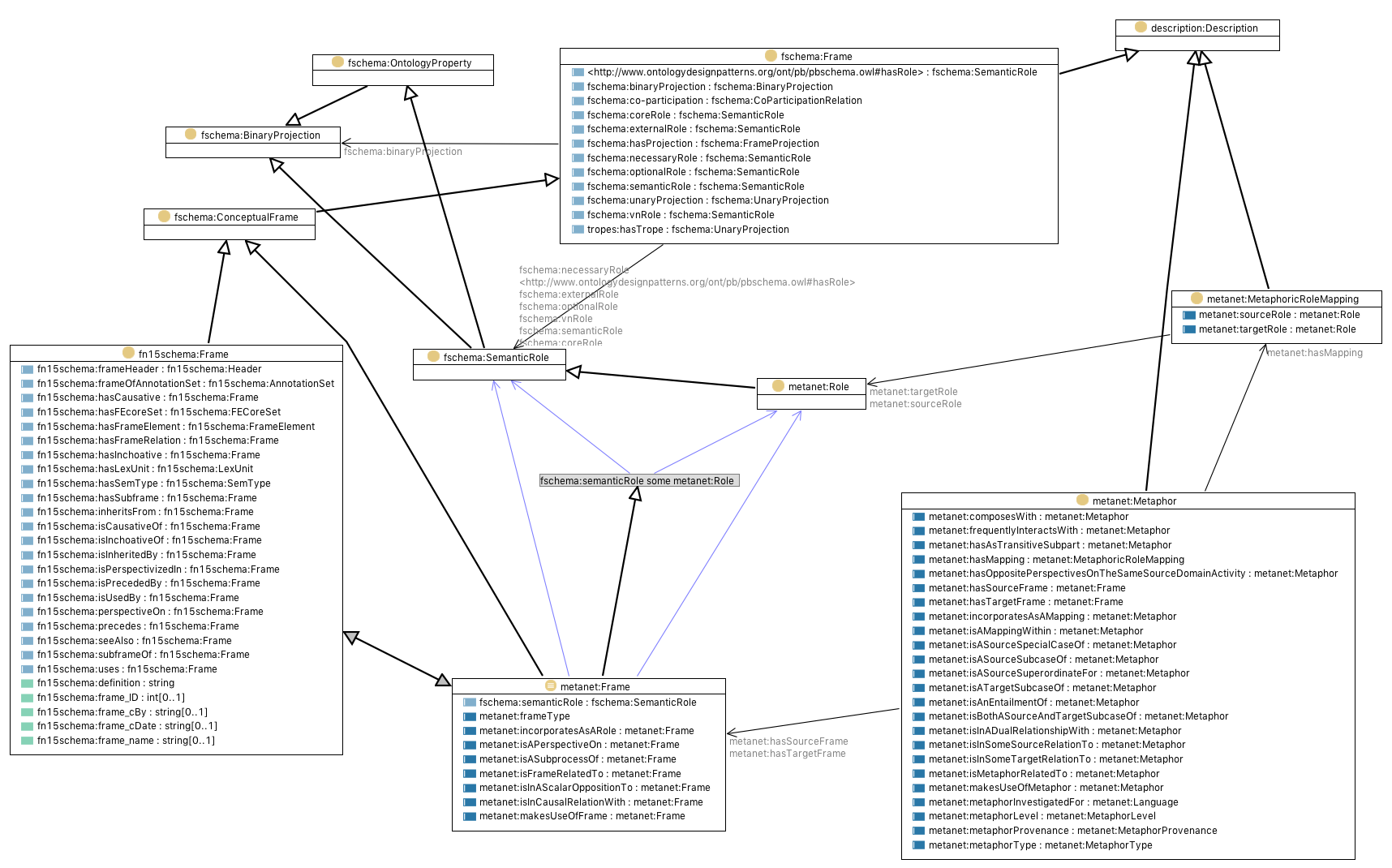}
 \caption{A class-diagram profile for the OWL axiomatisation of Amnestic Forgery.}
 \label{amnesticforgery}
\end{figure}

A summary diagram of the axiomatisation for Amnestic Forgery is showed in Fig. \ref{amnesticforgery}. The diagram uses a UML-class-diagram-oriented profile to sketch the core axioms for the \texttt{Metaphor} class,  shown either as ``attributes'' within  class boxes, or as either ``associations'' or  ``generalisations'' (subsumption) across class boxes. The diagram summarizes the reuse of the \texttt{Description} class from D\&S, which subsumes the \texttt{Metaphor},  \texttt{Frame}, and \texttt{MetaphoricRoleMapping} classes. A hierarchy of frame and role notions exemplify Framester schema alignements, and the treatment of semantic roles as both binary projection of frames, and OWL properties (binary relations). Association-like edges derive from either domain or range restrictions in the OWL encoding of Amnestic Forgery, or from existential restrictions. Please refer to the OWL file for the full axiomatization, including imports, alignments, disjointness, and documentation axioms.\footnote{\url{http://www.ontologydesignpatterns.org/ont/metanet/metanetschema.ttl}, use the same path for typing and data files: /metanettypes.ttl, /metanetdata.ttl. All material is also available on the GitHub site mentioned in Sect. \ref{conc}.}. 

In order to take full advantage of Framester, alignment of metaphors and frames to other resources such as FrameNet and WordNet is necessary. However, only about 25\% of MetaNet frames are aligned to FrameNet frames, therefore an alignment completion is needed. We have started this activity, which will be finalised by means of crowdsourcing methods. An example shows the non-triviality of this completion. The metaphor ABUSIVE\_POLITICAL\_LEADERS\_ARE\_PHYSICAL\_BULLIES could be aligned by using the alignments already existing in Framester, for example, \emph{abusive}, \emph{(political) leader}, and \emph{bully} are all present in WordNet, and are aligned to the following FrameNet frames (as unary projections) respectively: \texttt{Abusing}, \texttt{Leadership}, and \texttt{Manipulate\_into\_doing}. However, some MetaNet frames can only be aligned to FrameNet if we consider them as frame compositions: \texttt{Abusing}+\texttt{Leadership}, while others, like \texttt{Manipulate\_into\_doing}, require a specialisation to a further feature (here: physical). The alignment completion activity becomes then a discovery activity, where new frames can be proposed, and defined as compositions of existing frames.

\section{Generating new metaphors with MetaNet and Framester}
\label{generating}
In order to prove the advantages of having a large and formally rigorous knowledge base, we report here a query to Framester extended with Amnestic Forgery and MetaNet data. Given a MetaNet metaphor, the query is able to generate hundreds of novel intensional metaphors\footnote{A web application is being implemented to use this query graphically and with natural language generation methods to enhance user experience.}. Example results include the eponymous Amnestic Forgery as a linguistic rendering of the FORGERY\_IS\_AMNESIA metaphor, which appears to be actually novel: no realisations can be found e.g. on the Web (based on Google searching).
A second query type could be conceived in order to search for MetaNet metaphors that explain a metaphoric expression, such as \emph{bright coach}, taken from a large repository of metaphoric adjective-noun phrases\footnote{\url{http://pages.ucsd.edu/~e4gutier/m4p/AN-phrase-annotations.csv}}. 

The queries, which can be tested on the Framester SPARQL Endpoint\footnote{\url{http://etna.istc.cnr.it/framester2/sparql}}, as announced earlier only use the adjective-noun phrase construction, and their related senses and frames in Framester.

\begin{verbatim}
prefix metanet: <https://w3id.org/framester/metanet/schema/>
prefix framedata: <https://w3id.org/framester/metanet/frames/>
prefix metaphordata: <https://w3id.org/framester/metanet/metaphors/>
SELECT DISTINCT ?ssyn ?tsyn
WHERE {
	metaphordata:CRIME_IS_A_DISEASE metanet:hasSourceFrame ?s ; 
    		metanet:hasTargetFrame ?t .
	?s skos:closeMatch ?fns . ?fns a fn15schema:Frame .
	?t skos:closeMatch ?fnt . ?fnt a fn15schema:Frame .
	?fns skos:closeMatch ?ssyn .
	?fnt skos:closeMatch ?tsyn .
	{?ssyn a wn30schema:AdjectiveSynset}
	UNION 
	{?ssyn a wn30schema:AdjectiveSatelliteSynset}
	?tsyn a wn30schema:NounSynset }
\end{verbatim}

\section{Referential aspects of metaphorical mappings: a case for quasi-truth}
\label{referential}

As anticipated in Sect. \ref{intro}, we want to represent not only conceptual metaphors as frame mappings, but also the factual knowledge that is possibly affected by the those mappings. Does metaphor involve an actual referential movement in the domain of discourse, or is it just a ``false movement'', a sort of distorting lens that does not affect referential aspects at all? In the second case, the ontological relevance of metaphors would be confined to intensional aspects impacting cognitive or linguistic worlds, as Davidson would have probably subscribed due to his belief in purely linguistic relevance of metaphors \cite{doi:10.1086/447971}. In the first case, we should admit that frame mapping has a factual import.

Let's consider the trailing example: the CRIME IS A DISEASE metaphor as a frame mapping. Intensionally, it is pretty clear that after applying the metaphor, some roles from the \texttt{Crime} frame accept values from the \texttt{Disease} frame. However, what is happening extensionally? What are corresponding metaphoric situations like? 

Following Amnestic Forgery, a metaphoric situation is the counterpart of a metaphoric description, where entities from two frame occurrences are blended when playing the particular role mappings enabled by the metaphor. For example, since the CRIME IS A DISEASE axioms dictate that the \texttt{criminal\_activity} role from \texttt{Crime} is swapped for the \texttt{disease} role from \texttt{Disease}, and the \texttt{victim} role from \texttt{Crime} is swapped for the \texttt{patient} role from \texttt{Disease}, the denotation of sentence \ref{corruption}:
\begin{equation} \label{corruption}
\text{\parbox{.85\textwidth}{Corruption has infected our community}}
\end{equation}
would not be assigned within the literal domains of discourse. 

The literal assignment would work as follows: 
\begin{itemize}
\item a specific series of corruption events $ce$ is the value for the \texttt{criminal\_activity} role from a \texttt{Crime} situation $cs$: $CA(cs,ce)$
\item a specific series of infection events $ie$ is the value for the \texttt{disease} role from a \texttt{Disease} situation $ds$: $D(ds,ie)$
\item a specific community $com$ is the value for the \texttt{victim} role from a \texttt{Crime} situation $ds$: $V(cs,com)$
\end{itemize}

When the metaphor is activated, the mapping produces the following blended frame occurrence:
\begin{itemize}
\item a specific series of corruption events $ce$ is the value for the \texttt{disease} role from a \texttt{Crime+Disease} situation $cds$: $D(cds,ce)$
\item a specific series of infection events $ie$ is the value for the \texttt{disease} role from a \texttt{Crime+Disease} situation $cds$: $D(cds,ie)$
\item a specific community $com$ is the value for the \texttt{patient} role from a \texttt{Crime+Disease} situation $cds$: $P(cds,com)$
\end{itemize}

What happened? In this metaphor occurrence, once the metaphorical mapping generates the blended frame, the \texttt{criminal\_activity} role value either (a) results to be the same thing as the \texttt{disease} role value: $ce = ie$, or (b) melts into a new hybrid entity $ce+ie$. That hybrid entity is a strange beast, but in formal ontology we are accustomed to such things: aggregates, amalgams, qua-entities, etc.

The actual problem is if our design could satisfy application requirements by leaning to an economic commitment to our domain of interpretation $\Delta$: $ce=ie \in \Delta$, or else if we commit to a multiplicative design style, which accepts commitments also to a new entity $ice \in \Delta$, generated via metaphor.

The decision is not trivial. On one hand, we can safely claim that any physical substrate of entities involved in the blended frame are not changed by the metaphor, and that on the contrary, the space of cognitive entities is most probably affected. On the other hand, it is not obvious at all what are the social entities possibly involved in the blending.

A paradigmatic case is that of fake news, and more specifically of what we call \emph{quasi-true facts}, i.e. sentences that distort facts in a way that make them not really false, but ``alternative'' to (supposedly true) ones.\footnote{\emph{Alternative facts} are seriously used as a category in communication science talk, cf. \cite{alternativefacts}.} Spin doctors speech heavily adopts quasi-truth in order to obtain the best cognitive impact on citizens \cite{politicalmind}, e.g. when one political party narrates socially-relevant criminal cases by using the CRIME IS A DISEASE metaphor (in this case, typical entailment includes that that crime spreads like disease in human societies, and its etiology must be destroyed), while another uses the CRIME IS A PHYSIOLOGICAL PROCESS metaphor (in this case, the typical entailment is that crime is organic to human societies). 

When addressing the referential aspects of quasi-truth, the physical substrate of facts does not change, the cognitive impact is high (and can even lead to massive social change), while social reality still needs an appropriate characterisation: as a social entity, are corruption events -- seen like disease spreading -- a new entity in $\Delta$?

We do not propose a preferred design pattern for the social ontology of metaphors, since we'd rather defer conclusions after an ongoing empirical study. However, we do hope that the problems at hand are sufficiently clear after the application of Amnestic Forgery.\footnote{Funnily enough,  after naming the new ontology, the authors realised that Amnestic Forgery is a good literal description for  quasi-truth either.}

\section{Related work}
\label{related}
Occurrence of a metaphor in language is very frequent. It is not only what the meaning of a word is used in certain context, it is also deeply embedded in the fact that the words used in different domains are re-used differently in another context based on the knowledge a particular word or lexical unit reflects. The reference theory is Conceptual Metaphor Theory (CMT) proposed by Lakoff and Johnson in \cite{LakoffJohnson80}. 

Despite computational work started many years ago \cite{regier}, the computational metaphor literature is rather small, cf. \cite{DBLP:journals/coling/VealeSKHS18} for a survey. Recently the problem seems to have caught attention though. Computational applications nowadays include \emph{(i)} Machine Translation: the metaphors vary across cultures and languages, \emph{(ii)} Sentiment Analysis: Linguistic units may look positive or negative, but polarity could be inverted because of irony or figurative speech, \emph{(iii)} Information retrieval: faulty information may be found in textual data without properly attaching meaning to metaphorical phrases, \emph{(iv)} Computational Creativity: reverse engineering human ability to generate appropriate metaphors is key to computational creativity. 
Following these lines, three major tasks can be singled out: Metaphor Detection, Interpretation, and Generation. Where appropriate, we compare existing work to the ontology and themes described in this paper. 

\paragraph{Metaphor Detection.} Over the past few years there has been a rise in the development of statistical methods for detecting metaphors. Many of these techniques take advantage of vector-space models, and perform a binary classification of metaphorical vs. literal occurrences in text \cite{Utsumi2006, TsvetkovBGND14}, which may come from compositions of word pairs such as ``sweet'' and ``person'', where \emph{sweet} is only metaphorical when composed in phrases with words that do not denote tastable entities, such as ``person''. \cite{GutierrezSMB16} proposes a Compositional Distributional Semantic Model (CDSM), which generates a vector representation of the phrases. \cite{BaroniZ10} also introduces a framework based on CDSM, targeting adjective-noun constructions. In these cases, the meaning of the phrase is derived by composing the representations of adjectives and nouns. 
As a contrast, \cite{coling-ShutovaSK10} uses clustering techniques over nouns and verbs to perform metaphor identification. It takes manually annotated metaphors as a starting point, and then, based on syntactic similarity, detects a large number of metaphors from a corpus. \cite{KlebanovLGSF16} uses binary classification of the verbs into metaphorical or literal using semantic classes of the verbs such as grammatical, resource-based, or distributional. While in \cite{W17-4906} the authors propose a two step approach for detecting if the an adjective-noun pair is a metaphor or a literal. They use pre-trained word vectors of the AN-pairs as input vectors and then propose a neural network for composing AN phrases. 
Finally, \cite{ChenLLH17} uses eventive information in detecting metaphors in Chinese.
Metaphor detection can be nicely paired with metaphor-oriented knowledge graph processing. For example, we plan to pre-process  corpora with a model trained with combinatorially-generated metaphors, in order to detect if such metaphors are used, and in what context, thus generating additional knowledge.

\paragraph{Metaphor Interpretation.} Metaphor interpretation requires complex analogical comparison and inferencing, because it  performs cross-domain projection of knowledge i.e., projecting the knowledge from one domain to another. One way to assign interpretations to metaphors relies on MetaNet \cite{W15-1405}. Each metaphor defined in MetaNet is manually encoded, and is connected to a combination of linguistic frames, often aligned to those available in FrameNet \cite{shortBaker1998}. A detailed survey about metaphor processing systems is \cite{coling-Shutova15}.
As far as our ontology is concerned, we intend to enrich knowledge extraction pipelines such as FRED \cite{DBLP:journals/semweb/GangemiPRNDM17} with metaphoric sensitivity, thus contributing to automated metaphor interpretation.

\paragraph{Metaphor and Blending Generation.} Tony Veale has devoted substantial effort to make automated metaphor generation a reality, cf. \cite{veale} for a summary of his recent attempts to make it an creative agent on the social media.
The most advanced computational framework for automated conceptual blending is Eppe et al. \cite{DBLP:journals/ai/EppeMCKSPK18}. It is able to implement most of the nuances of the theory, but does not make use of knowledge graphs or public ontologies (they only mention future work with the OntoHub repository). Amnestic Forgery is obviously reusable by them in order to experiment with new case studies involving metaphorical blending, as well as by reusing the whole Framester as background knowledge.

Interesting formal ontology work has also been conducted with reference to cognitive conceptual spaces \cite{DBLP:journals/semweb/RaubalA10}, or description logics \cite{DBLP:conf/esslli/SchorlemmerCP16}, but none of these works attempts to build a cognitively valid metaphor ontology that can be also exploited empirically in the current huge knowledge graphs that started populating the Web. 

The work reported in this paper is to our knowledge the first attempt to position metaphor and blending theories within an open large graph that can be used to test or extend existing theories, as well as classification, interpretation, or generation algorithms.

\section{Conclusions}
\label{conc}
We have presented Amnestic Forgery, an ontology that enables the integration of Conceptual Metaphor Theory data into large knowledge graphs. The ontology is designed starting from the D\&S ontology pattern framework, and inherits the Framester schema, which already integrates multiple linguistic and factual data and schemas, besides  some foundational ontologies.

The resource is intended to be used as background knowledge for empirical formal research on metaphor phenomena in discourse.
Amnestic Forgery can be queried or downloaded through the Framester SPARQL Endpoint or its GitHub repository\footnote{\url{https://github.com/alammehwish/AmnesticForgery}}.

The benefits of integrating a large metaphor dataset into the Framester linguistic-factual graph may spread to any research work on metaphor: metaphoric sentence classification, metaphor generation, as well as empirical research on cognitive or formal theories of metaphor.

The paper also includes references to the current computational research on metaphors, describes a simple SPARQL-based method to generate novel metaphors, and discusses the challenging referential problems of metaphorically-induced situations in social reality, paving the way to a theory of quasi-truth.

Ongoing and future work bears multiple directions: using deep learning techniques to both detect metaphorical sentences in text, and automatically enriching metaphoric knowledge graphs; deepening the axiomatisation of Amnestic Forgery, and empirically testing it on large focused corpora; integrating Amnestic Forgery with computational blending and creativity platforms.

\bibliographystyle{plain}
\bibliography{fois2018}

\end{document}